\documentclass[twoside,11pt]{article}

\usepackage{blindtext}

\usepackage[abbrvbib, preprint]{jmlr2e}
\usepackage[utf8]{inputenc} % allow utf-8 input
\usepackage[T1]{fontenc}    % use 8-bit T1 fonts
\usepackage{hyperref}       % hyperlinks
\usepackage{url}            % simple URL typesetting
\usepackage{booktabs}       % professional-quality tables
\usepackage{amsfonts}       % blackboard math symbols
\usepackage{nicefrac}       % compact symbols for 1/2, etc.
\usepackage{microtype}      % microtypography
\usepackage{xcolor}         % colors
\usepackage{graphicx}
\usepackage{amsmath}
\usepackage{natbib}
\usepackage{subfigure}
\usepackage{comment}
\usepackage{array}
\usepackage{mdframed}
\usepackage{tabularx}
\usepackage{ragged2e}
\usepackage{mathtools}
\newcolumntype{C}[1]{>{\Centering}p{#1}}
\newcolumntype{L}{>{\raggedright}X}

\ShortHeadings{}{}
\firstpageno{1}

\begin{document}

\title{Rethinking recidivism through a causal lens}

\author{\name Vik Shirvaikar \email vik.shirvaikar@spc.ox.ac.uk \\
       \addr Department of Statistics \\
        University of Oxford \\
        Oxford, United Kingdom
       \AND
       \name Choudur K. Lakshminarayan \email lchoudur@stevens.edu \\
       \addr School of Business \\
        Stevens Institute of Technology \\
        Hoboken, New Jersey, United States \\}

%\editor{My editor}

\maketitle

\begin{abstract}%   <- trailing '%' for backward compatibility of .sty file
Predictive modeling of criminal recidivism, or whether people will re-offend in the future, has a long and contentious history. Modern causal inference methods allow us to move beyond prediction and target the ``treatment effect'' of a specific intervention on an outcome in an observational dataset. In this paper, we look specifically at the effect of incarceration (prison time) on recidivism, using a well-known dataset from North Carolina. Two popular causal methods for addressing confounding bias are explained and demonstrated: directed acyclic graph (DAG) adjustment and double machine learning (DML), including a sensitivity analysis for unobserved confounders. We find that incarceration has a detrimental effect on recidivism, i.e., longer prison sentences make it more likely that individuals will re-offend after release, although this conclusion should not be generalized beyond the scope of our data. We hope that this case study can inform future applications of causal inference to criminal justice analysis.
\end{abstract}

%\begin{keywords}
%  causal inference, recidivism
%\end{keywords}

\section{Introduction}

Algorithms for recidivism risk assessment are widely used in criminal justice adjudications across the world, including bail hearings, parole reviews, and sentencing \citep{barnes2012classifying, tollenaar_which_2013, desmarais_performance_2016}. This practice garnered significant public attention in 2016 following ProPublica's investigation of the privately-developed COMPAS program, which has used an elaborate 137-item questionnaire to evaluate over a million defendants' risk of recidivism since 1998 \citep{angwin2016machine}. The study and ensuing debate mainly highlighted racial disparities in the program's risk scores, though follow-up research showed that ProPublica may have misestimated the extent of algorithmic bias \citep{chouldechova_fair_2016, rudin_age_2020}. 

However, regardless of race, it was found that COMPAS's predictions tended to be inaccurate. For the binary classification task of whether offenders would recidivate, the program's benchmarked accuracy was 65.4$\%$, but with the same data, a two-variable regression based only on age and prior convictions achieved an accuracy of 66.8\%, casting ``significant doubt on the entire effort of algorithmic recidivism prediction'' \citep{dressel_accuracy_2018}.  Several other studies have similarly demonstrated that due to inherent noise in the recidivism domain, sophisticated machine learning techniques often fail to outperform simple regression models \citep{liu_comparison_2011, tollenaar_which_2013, wang_pursuit_2023}.

As a result, though existing recidivism literature focuses almost exclusively on predictive modeling, we turn our attention to \emph{causal inference} methods, which enable us to conduct a more nuanced analysis of the social determinants of recidivism. Approaches such as potential outcomes \citep{rubin_estimating_1974} and directed acyclic graph (DAG) analysis \citep{pearl_causality_2000} allow for causal ``treatment effects'' to be estimated from observational data by adjusting for the confounding effects of underlying covariates. To demonstrate this, we look specifically at the effect of incarceration on recidivism (under the interpretation of prison time as a ``treatment'' imposed by the criminal justice system). Prior studies have assessed this relationship, but primarily via traditional econometric techniques, such as instrumental variables and discontinuity analysis, that rely on the identification of a suitable natural experiment \citep{rose_how_2021, loeffler_impact_2022, stevenson_conviction_2023}. 

The fundamental problem in observational causal analysis is that the treatment mechanism may be confounded. In other words, since the treatment cannot be administered randomly (as in a controlled clinical trial), certain observed or unobserved factors may influence both individuals' outcomes and their propensities of receiving treatment. When considering the causal effect of incarceration on recidivism, this issue is especially prevalent: individuals who received longer sentences have likely committed more numerous or more serious crimes, potentially indicating a higher propensity to recidivate following release. A naive hypothesis test or regression analysis would therefore misstate the effects of incarceration. Fortunately, adjusting for these confounding effects can be possible if we have access to criminal history information. We demonstrate using two popular approaches from the causal inference literature, outlined below and described in detail in Section 3.

\begin{figure}[h]
\centering
\includegraphics[width=.3\textwidth]{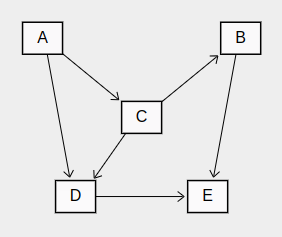}
\caption{Example of directed acyclic graph (DAG)}
\label{fig:sample}
\end{figure}

The first approach is based on directed acyclic graph (DAG) adjustment, as promoted by Judea \citet{pearl_causality_2000} and collaborators. In this framework, a causal graph that encodes the relevant relationships between variables is first elicited. Figure \ref{fig:sample} displays an example DAG, in which variable A causes C and D; B causes E; and so on. Graph elicitation is informed by domain knowledge as well as the use of causal discovery methods, which posit and verify possible edges in the graph by analyzing conditional independence relationships in the data. Next, the edge of interest (in our case, between prison time and recidivism) is identified and isolated by blocking all other non-causal paths linking the treatment and outcome. Specifically, this is accomplished through the use of Pearl's backdoor criterion, a set of rules for determining a valid set of adjustment variables. Controlling for these variables, such as in a generalized linear model (GLM) or Cox proportional hazards survival model, then refines an initial estimate of association to yield a causal effect.

The second approach is double/debiased machine learning (DML), as presented by \citet{chernozhukov_doubledebiased_2018}. In this framework, the treatment and outcome are first each predicted separately as functions of the covariates. Notably, this can be done through a variety of flexible machine learning methods, such as random forests or regularized GLMs, allowing for complex nonlinear relationships. The resulting estimates allow us to calculate each individual's ``residual'' (or error) in both treatment and outcome, obtained by subtracting the predicted model values from the observed data. In our context, these values represent how much more or less prison time an individual received than their expected sentence (based solely on their covariates); and, similarly, how likely or unlikely their recidivism outcome was (again, based solely on their covariates). The adjusted treatment effect is then calculated as a function of these residual values. Consistent estimation is guaranteed via Neyman orthogonality, which ensures that the final effect estimate is valid even if the separate treatment and outcome models are slightly misspecified. \citet{chernozhukov_long_2022} additionally provide a technique for sensitivity analysis within the DML framework, allowing us to determine how strong an unobserved confounder would have to be in order to explain away the identified causal effect.

For demonstration, we use a dataset from North Carolina, originally collected in 1978 and 1980 \citep{schmidt_data_1989}. This particular dataset has been widely covered in the literature and is well-established as a case study for recidivism analysis \citep{schmidt_split_1989, chung_survival_1991, palocsay_predicting_2000}. Due to its age and limited geographic scope, our conclusions should not be interpreted as a definite or generalizable statement about the causes of recidivism. Rather, our goal is simply to demonstrate a methodological pipeline that can then be applied to other case studies of interest. In this particular dataset, we consistently find that incarceration has a positive causal effect on recidivism, i.e., longer prison sentences increase the probability that prisoners will re-offend in the future. This effect can be interpreted as causal in the sense that it accounts and controls for underlying differences in criminal history, crime type, etc.

The remainder of this paper is organized as follows. Section 2 outlines the history and background of quantitative recidivism analysis. Section 3 describes our methods and Section 4 contains our results. Section 5 discusses and concludes.

\section{Background and Related Work}

In quantitative recidivism analysis, available data generally includes demographic information (such as age, gender, marital status, etc.) and criminal history records (such as crime type, prior offenses, etc.). It may also include other socioeconomic factors, such as education, employment status, and so on. These datasets typically define a time period following release from prison (for example, two to four years) and indicate recidivism within that period as a binary variable. In this section, we give a brief overview of existing efforts in the area.

\subsection{Predictive Modeling}

When faced with covariates and a binary outcome, the immediate instinct of most quantitative researchers is to build a predictive model. Recidivism prediction has been an active research area for decades, as social scientists recognized its potential value in enhancing public safety and informing more effective rehabilitation policies. However, by the 1980s, a review of the field reveals a general recognition of the inherent difficulty in accurately predicting recidivism, due to the wide array of (often unmeasurable) situational and personal factors that can affect criminal behavior. 

\citet{schmidt_book_1988} provide an instructive comparison to a widely-studied economic problem: the literature on determinants of individual wages is robust, with precise measurements available for both outcomes and key predictors (education, job title, experience, etc.), but researchers are still generally unable to explain even 50\% of the variation in wages (i.e. $R^2 < 0.5$). In comparison, recidivism is a much noisier domain:

\vspace{0.2cm}
\noindent \begin{minipage}{\textwidth}
    \textit{"We expect that criminal activity has a much larger random element than the wage rate and that we may never be able to explain more than 30\% or so of the variation in typical variables representing criminality, even under ideal conditions."}
    \hfill --- \citet{schmidt_book_1988}
\end{minipage}
\vspace{0.2cm}

For the leading recidivism models of the time, \citet{gottfredson_prediction_1987} finds that “the proportion of the outcome variance explained rarely exceeds 15 to 20\% and is most frequently lower”. Ultimately, explaining variance is not strictly necessary if these models can still predict well, but \citet{lerner_thoughts_1990} show that they return false positive and false negative rates regularly exceeding 30\% to 50\%, making it ``very hard to justify differential treatment of individuals'' based on their predictions. 

A common refrain at this point was that the continuing development of more powerful predictive methods would eventually solve this problem; however, the application of modern machine learning (ML) to recidivism prediction has yielded mixed results at best. \citet{caulkins_predicting_1996} found that neural networks provided no advantage over conventional methods, while \citet{palocsay_predicting_2000} found some benefit over logistic regression, but at the cost of significantly less interpretability. Later studies from \citet{liu_comparison_2011} and \citet{tollenaar_which_2013} similarly found effectively equivalent performance across a range of methods, including logistic regression, classification trees, linear discriminant analysis, and neural networks. 

\citet{zeng_interpretable_2017} note that when several different models produce roughly equivalent predictive accuracy, referred to as the ``Rashomon'' effect by \citet{breiman_statistical_2001}, the goal should shift to finding those that are fair and interpretable while remaining accurate. Several recent efforts have therefore focused on this aim, with promising results. \citet{wang_pursuit_2023} provide a review, showing that certain simpler models (often using linear or tree-based representations) can provide similar performance to black-box ML, but with better interpretability and stronger guarantees of fairness. The design and implementation of these predictive tools remains an open research question of considerable interest.

\subsection{Survival Analysis}

A related strain of research aims to reframe recidivism as a survival analysis problem. The previously discussed predictive models generally indicate recidivism within a specified time period as a binary variable. Survival models aim to predict not only whether an event will happen but also when. These models are commonly applied in medical settings to predict the time until an adverse medical outcome given an individual's health status, and in industrial settings (where they are referred to as ``reliability'' or ``failure-time models'') to predict the time until a mechanical system fails.

In our case, the clock starts upon release from prison, and ends (if it ends at all) upon recidivism. Survival analysis therefore aims to model how each covariate will either increase or decrease the amount of time until an individual recidivates. \citet{schmidt_book_1988} note that this provides significant advantages over binary prediction. It utilizes the complete information contained in each individual's time until recidivism; it acknowledges that data collection has to be stopped or censored at some given point; and it allows us to model the probability of recidivism within any given future time interval, not just the pre-specified binary cutoff.

An immediate concern is that, as evidenced by the medical and industrial use cases, survival analysis implicitly assumes that the endpoint event must eventually happen, but some released offenders will never actually commit another crime. \citet{maltz_mathematics_1977} accordingly find that representing recidivism as a ``split-population'' survival model can improve performance, where the baseline probability of recidivism is assumed to be less than one and is modeled separately from its conditional timing. The original studies using our North Carolina dataset also apply a split-population model \citep{schmidt_split_1989, chung_survival_1991}. In recent applications, binary prediction continues to be the most common approach for recidivism data, but survival models are also applied on occasion \citep{monnery_determinants_2013, peters_parolee_2015, tollenaar_optimizing_2019}.

\subsection{From Prediction to Causality}

Considering the difficulties inherent in accurate prediction, it was quickly recognized that quantitative methods could alternatively be used to analyze individual covariates as determinants of recidivism:

\vspace{0.2cm}
\noindent \begin{minipage}{\textwidth}
    \textit{"When the sample size is large, a model which explains only 10\% of the variation in the dependent variable can yield statistically significant statements about the effect of explanatory variables. Such a model can be useful from the point of view of both theory and policy."} \\
    \textcolor{white}{fill} \hfill --- \citet{schmidt_book_1988}
\end{minipage}
\vspace{0.2cm}

However, ``statistical significance'' generally only indicates correlation, limiting the interpretation of any such results. Recent methods and developments have therefore attempted to bridge the gap from correlation to causation. \citet{samii_retrospective_2016} applied retrospective potential outcome methods to assess the differential effects of various policy interventions on recidivism in Colombia, while several studies have also used potential outcomes and counterfactual approaches to assess the fairness of risk assessment algorithms \citep{khademi_algorithmic_2019, khademi_fairness_2019, berk_improving_2023}.

These approaches apply modern causal frameworks, but are focused on big-picture issues such as overall algorithmic bias and discrimination. However, current causal literature, particularly from the biomedical and clinical trial arenas, provides a wealth of methods to identify and target the effect of an individual treatment on an outcome. These methods, such as the directed acyclic graph (DAG) and double machine learning (DML) approaches we apply, are applicable to observational data not collected in a randomized study, making them compelling candidates for recidivism analysis.

As our ``treatment'' variable of interest, we consider the effect of incarceration (prison time) on future recidivism. Of the information available in recidivism data, prison time most closely resembles an actual intervention or treatment, since its overall level (initially determined at sentencing and potentially modified later through parole or early release) is ``administered'' by the criminal justice system. Prior studies have considered this question, but largely with econometric methods that require a natural experiment or exogenous source of random variation to be found. Examples include instrumental variable approaches based on the random assignment of judges to cases \citep{rose_how_2021, loeffler_impact_2022} and discontinuity approaches based on specific quirks found in sentencing guidelines \citep{loeffler_impact_2022, stevenson_conviction_2023}.

\section{Methods}

To analyze our question through a causal lens, we present two approaches from the causal literature. The first applies causal discovery to infer a graph structure from the underlying data, then uses ``traditional'' graph adjustment methods, specifically the backdoor criterion of \citet{pearl_causality_2000}, to identify the causal effect of interest. The second applies the modern approach of double/debiased machine learning (DML), as presented by \citet{chernozhukov_doubledebiased_2018}, to learn separate models for the treatment and outcome, then uses these models to ``debias'' the effect of confounding variables. Code and data are available at \url{https://github.com/vshirvaikar/recidivism}.

\subsection{Causal Discovery and Adjustment}

In a directed acyclic graph (DAG) model, each node represents a variable, while each edge is an arrow indicating a (causal) relationship from the variable at its tail to the one at its head. Importantly, any DAG can be equivalently represented as a set of implied conditional independences between variables. For example, in Figure \ref{fig:sample}, A is independent of B given C $(A \bot B \mid C)$; B is independent of D given C $(B \bot D \mid C)$; and so on.

Causal discovery methods aim to infer a causal graph structure from a set of observed data \citep{glymour_review_2019}. This approach was first used in the context of ``Bayesian networks'' designed to learn high-dimensional models of gene expression from microarray data \citep{spirtes_constructing_2000}. The two original methods, PC (named after its founders' initials) and Fast Causal Inference (FCI), use a series of conditional independence checks and rules to determine possible relationships between the given variables. These methods are referred to as ``constraint-based'' since they systematically find constraints on the set of valid graphs. Later approaches based on functional causal models aim to relax some of the distributional assumptions made by constraint-based algorithms \citep{shimizu_linear_2006}, while modern machine learning approaches extend this to a generalized continuous optimization problem \citep{zheng_dags_2018}.

To facilitate simplicity and explainability in our application, we revert to the classical PC algorithm. However, the standard PC conditional independence tests are either the Gaussian partial correlation test (for continuous variables) or the $G^2$ test (for binary variables), neither of which can be applied to mixed data with both types. \citet{tsagris_constraint-based_2018} present a conditional independence test for mixed data, which relies on likelihood-ratio tests between nested models for the target variable. We implement this approach through the R package \textbf{MXM} in order to apply the PC algorithm for our use case \citep{biza_mxm_2022}.

With a causal graph in hand, we then apply the methods of \citet{pearl_causality_2000} to adjust our graph and identify the causal effect of interest. The backdoor criterion says that, to adjust for the effect of a treatment $T$ on an outcome $Y$, we must find some set of variables $Z$ such that (1) no node in $Z$ is a descendant of $T$ and (2) $Z$ blocks every path between $T$ and $Y$ that contains an arrow into $T$. If we now fit a model for $Y$ that includes both $T$ and the adjustment set $Z$, we can interpret the coefficient for $T$ as an unbiased estimate of the true treatment effect. To facilitate explainability, we first demonstrate this with a simple logistic regression model. 

As noted in Section 2.2, there are compelling reasons to model recidivism as a time-to-event or survival problem, so we perform a Cox proportional hazards regression as well \citep{cox_regression_1972}. Cox regression adds an additional layer of nuance by not merely focusing on whether an event occurs but also on when it happens. This is represented through a hazard function
\begin{equation}
    h(t \mid X) = h_0(t) e^{\beta_1 x_1 + ... + \beta_p x_p}
\end{equation}
which indicates the instantaneous risk of the event occurring at time $t$ (note that this is distinct from the treatment $T$) as a function of the covariates $X$. The baseline function $h_0(t)$ is the hazard when all covariates are zero. The overall distribution of survival times is then given by the survival function
\begin{equation}
    S(t) = e^{-\int_0^t h(u) du}
\end{equation}
which represents the probability of the event not occurring by time $t$. In practice, for any two individuals, the ratio of their hazard functions depends only on their covariates $X$ and not the baseline hazard $h_0(t)$; this proportional hazards assumption allows us to estimate how the covariates increase or decrease the risk of the event without specifiying the exact form of the risk over time.

\subsection{Double/Debiased Machine Learning}

Double/debiased machine learning (DML) is a popular and flexible approach that allows for modern machine learning (ML) methods to be applied to the estimation of causal effects \citep{chernozhukov_doubledebiased_2018}. ML algorithms, while excellent at yielding accurate predictions, can be inconsistent in identifying causal effects on outcomes. For example, consider the partially linear regression model
\begin{equation}
Y = T \theta_{0} + g_0(X) + \epsilon    
\end{equation} 
where the outcome $Y$ is linearly related to the treatment $T$ but has a more complex, potentially nonlinear relationship $g_0$ with high-dimensional covariates $X$. If we used an ML method to fit this model directly by minimizing some loss function, such as the mean-squared error, our estimate of the treatment effect $\hat{\theta}_0$ may be biased due to misestimation of the $\hat{g}_0$ function.

In order to refine our estimate of $\theta_0$, we therefore first estimate a separate propensity model $T = m_0(X) + \zeta$ that represents the treatment as a complex function $m_0$ of the covariates. Using the fitted $\hat{m}_0$ from this step, we can calculate $\hat{\zeta} = T - \hat{m}_0(X)$ as each individual's ``residual'' or error in treatment, i.e., how much more or less treatment he/she received than would have been expected based only on the covariates. From the initial step, we also have $\hat{\epsilon} = Y - \hat{g}_0(X)$ as the equivalent representation of each individual's ``residual'' or error in outcome. We use these separate models to formulate the debiased estimator 
\begin{equation}
\label{eq:double}
\hat{\theta}_0 = \left( \frac{1}{n} \sum_i \hat{\zeta}_i T_i \right)^{-1} \frac{1}{n} \sum_i \hat{\zeta}_i (Y_i - \hat{g}_0(X_i))
\end{equation}
which deconfounds both $T$ and $Y$, yielding a consistent estimate of the treatment effect. We implement this through the R package \textbf{DoubleML}, which provides several options for specification of the outcome and treatment models above \citep{bach_doubleml_2024}.

In the statistical literature, this framework falls under the umbrella of semi-parametric inference because our target parameter $\theta_0$ is finite-dimensional, but we have to estimate $\eta_0 = \{g_0, m_0\}$ as a separate infinite-dimensional nuisance parameter. The novelty of DML is in showing that the debiased estimator above is root-$n$ consistent, meaning that the estimation error reliably decreases at an optimal rate even if the nuisance parameters are misspecified. This is achieved via a proof based on Neyman orthogonality, a property which is mathematically represented by the moment condition
\begin{equation} 
\label{MC}
E[(\underbrace{(Y-E[Y|X])}_{\mathclap{\text{regression adjustment}}}-(T-E[T|X]) \; \theta_{0}) \underbrace{(T-E[T|X])}_{\mathclap{\text{propensity score adjustment}}}] = 0
\end{equation}
This condition debiases the influence of the confounders $X$ on the outcome $Y$ (regression adjustment) and treatment $T$ (propensity score adjustment) separately. The estimate of $\theta_{0}$ is then obtained by solving the empirical analog of Equation \ref{MC} via maximum likelihood estimation. Intuitively, Neyman orthogonality ensures that Equation \ref{eq:double} still reaches the optimum for the treatment effect $\theta_0$ in a small neighborhood around $\eta_0$, meaning the treatment and outcome functions $\{g_0, m_0\}$ can be slightly incorrect without affecting $\theta_0$. DML therefore offers a robust method for causal inference in complex settings that isolates the treatment effect from confounding.

The DML framework also extends to provide methods for sensitivity analysis. Specifically, \citet{chernozhukov_long_2022} derive novel bounds on the magnitude of omitted variable bias as a function of the explanatory power of unobserved confounders. This means we can determine how strong an unobserved confounder would have to be, in terms of both the outcome and treatment models, in order to nullify the estimated causal effect. We conduct this analysis and provide illustrative contour plots through the R package \textbf{dml.sensemakr} \citep{cinelli2023sensitivity}.

\section{Results}

In this section, we first discuss the dataset we will be using, and some important limitations with respect to scope and conclusions. We then present results from both approaches discussed previously.

\subsection{Data and Limitations}
\label{sec:data}

\begin{table}
\centering
\begin{tabular}{ccccccccccccccccc}
\toprule
& $y$ & $t_1$ & $t_2$ & $t_3$ & $x_1$ & $x_2$ & $x_3$ & $x_4$ & $x_5$ & $x_6$ & $x_7$ & $x_8$ & $x_9$ & $x_{10}$ & $x_{11}$ & $x_{12}$ \\
\midrule
Name & \parbox[t]{2mm}{{\rotatebox[origin=c]{90}{Recidivism}}} & 
\parbox[t]{2mm}{{\rotatebox[origin=c]{90}{Prison Time}}} & 
\parbox[t]{2mm}{{\rotatebox[origin=c]{90}{Follow-Up Time}}} & 
\parbox[t]{2mm}{{\rotatebox[origin=c]{90}{Return Time}}} &
\parbox[t]{2mm}{{\rotatebox[origin=c]{90}{Age}}} & 
\parbox[t]{2mm}{{\rotatebox[origin=c]{90}{Gender}}} & 
\parbox[t]{2mm}{{\rotatebox[origin=c]{90}{Race}}} &
\parbox[t]{2mm}{{\rotatebox[origin=c]{90}{Marital Status}}} & 
\parbox[t]{2mm}{{\rotatebox[origin=c]{90}{Felony Crime}}} & 
\parbox[t]{2mm}{{\rotatebox[origin=c]{90}{  Property Crime  }}} & 
\parbox[t]{2mm}{{\rotatebox[origin=c]{90}{  Personal Crime  }}} & 
\parbox[t]{2mm}{{\rotatebox[origin=c]{90}{Prior Offenses}}} & 
\parbox[t]{2mm}{{\rotatebox[origin=c]{90}{Parole Status}}} & 
\parbox[t]{2mm}{{\rotatebox[origin=c]{90}{Education}}} & 
\parbox[t]{2mm}{{\rotatebox[origin=c]{90}{Alcohol Use}}} & 
\parbox[t]{2mm}{{\rotatebox[origin=c]{90}{Hard Drug Use}}} \\
\midrule
A & 0 & 2.5 & 6 & N/A & 36.8 & 1 & 1 & 1 & 0 & 0 & 0 & 0 & 1 & 7 & 1 & 0 \\
B & 1 & 0.6 & 6.8 & 2.4 & 24.3 & 1 & 0 & 1 & 0 & 1 & 0 & 8 & 0 & 9 & 0 & 0 \\
C & 0 & 1.1 & 5.9 & N/A & 23.1 & 1 & 1 & 0 & 0 & 0 & 0 & 1 & 1 & 12 & 0 & 0 \\
\bottomrule
\end{tabular}
\caption{North Carolina dataset sample view}
\label{tab:NCsample}
\end{table}

The North Carolina dataset that we investigate in this study is one of several recidivism datasets available at the Inter-university Consortium for Political and Social Research (ICPSR) repository \citep{schmidt_data_1989, bradshaw_cross-validation_1989, gottfredson_criminal_1993}. Table \ref{tab:NCsample} displays a sample clip from the North Carolina data, with a mix of numerical and binary variables. Available fields include demographic information (age, gender, race, and marital status); criminal history information (crime degree and type, prior offenses, prison violations, and parole status); and other socioeconomic factors (education, alcohol use, and drug use). A data dictionary with the complete interpretation of each field can be found in Appendix \ref{app:NCdata}.

This dataset is well-known in the recidivism literature, having been studied several times previously, though largely from a predictive accuracy viewpoint \citep{schmidt_split_1989, chung_survival_1991, palocsay_predicting_2000}. We again emphasize that due to its restricted scope (a single U.S. state, within the years 1978 to 1980), our goal is only to demonstrate a promising approach for causal recidivism analysis. Our results should not be interpreted as policy recommendations or generalizable statements about the determinants of recidivism, especially in different regions or time periods.

To simplify interpretation, we limit our analysis to information from before the start of each individual's recorded prison sentence, and their subsequent incarceration and recidivism outcome. In particular, this means that we do not look at behavior while in prison or its resulting effect on parole status. In this dataset, it is noted that the prison time variable ($t_1$) indicates actual time served, not the length of the original sentence, meaning that defendants who qualified for parole or early release would have this value decreased accordingly. We consider this to still fit within the framework of incarceration as ``treatment'', but where the individual treatment amount can be dynamically altered (in this case, by a parole judge or board) if deemed appropriate. However, an objection is that paroled individuals could be less likely to recidivate due to the conditions of their parole, such as regular appointments or supervision. We therefore present each result for both (1) all individuals and (2) the subset of only individuals who received parole, making up 76\% of the data, to ensure robustness of conclusions. 

A related but distinct objection is that the impact of imprisonment could vary based on whether an individual's prior offenses were more or less violent. As a result, we additionally present each result for the mutually exclusive subsets of (3) those whose most recent offense was a felony and (4) those whose most recent offense was a misdemeanor, making up 37\% and 63\% of the data respectively, where the crime degree serves as our indicator of severity.

Finally, we address one of the central challenges in causal inference: the assumption of no unobserved confounders. Such confounders, or unseen fields that affect both incarceration and recidivism, could skew our findings, presenting a potentially insurmountable concern for social scientists or practitioners. For example, possibly relevant categories of data missing from our study include employment status and income; physical and mental health; or neighborhood and housing situation. However, the sensitivity analysis provided in Section \ref{sec:dmlresult} suggests that, in order to explain away our main causal effect, any unobserved confounders would need to improve the R-squared for both treatment and outcome by over 12\% (compared to a baseline R-squared of 24\% for treatment and 7\% for outcome). Our dataset already encompasses a broad spectrum of variables as outlined above, meaning that any confounders, including those mentioned above, are highly unlikely to be this strong even in combination. Ultimately, the utility of our analysis is inherently bound by the confines of the available data. However, we have strong reason to believe that this dataset captures, or at least approximates via proxy, a sufficient proportion of possible confounders. 

\subsection{Causal Discovery and Adjustment}

\begin{figure}
\centering
\includegraphics[width=.9\textwidth]{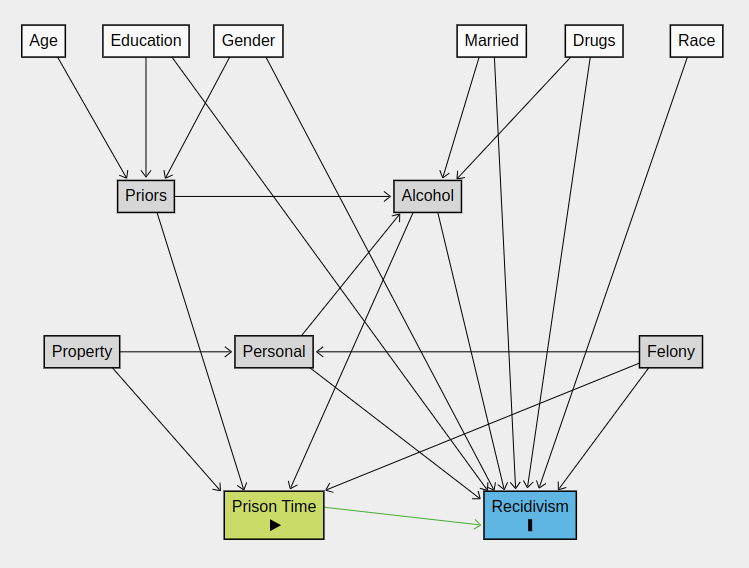}
\caption{Directed acyclic graph for North Carolina recidivism data, elicited using PC algorithm with mixed-data conditional independence testing}
\label{fig:main}
\end{figure}

Figure \ref{fig:main} displays the causal graph returned by the PC algorithm applied to our dataset, with the \textbf{MXM} package providing conditional independence tests for mixed data. At this stage, it is important to note that a causal graph only serves as an abstracted representation of reality for the purpose of facilitating further analysis, and that causal discovery is no substitute for domain expertise. 

In this case, however, we observe that the elicited graph largely aligns with what  might be expected given knowledge of the problem setting. Demographic variables such as age, race, and gender are generally at the top, causally prior to all other factors, with criminal history information such as prior offenses in the middle. The treatment (prison time) and outcome (recidivism) are at the bottom of the graph, with a complex web of factors affecting both, and prison time affecting recidivism. One peculiar feature is race, which is found to cause recidivism and nothing else. We speculate that this may indicate race serving as a proxy for other unmeasured social or economic factors; in any case, it does not affect the subsequent construction of an adjustment set.

We therefore advance to identification of this valid adjustment set via the backdoor criterion. In Figure \ref{fig:main}, an example of a valid backdoor adjustment set is priors, alcohol use, property crime, personal crime, and felony. This contains all available information on criminal history, including prior offenses and previous crime type and degree, along with alcohol use as a high-leverage predictor. In Table \ref{tab:logistic}, we present adjusted logistic regression results, with recidivism as a binary outcome, using this adjustment set. %As discussed in the previous section, we report results for (1) all defendants, (2) only those who received parole, (3) only those whose previous offense was a felony, and (4) only those whose previous offense was a misdemeanor.

We observe a positive treatment effect (the coefficient of prison time) in all four cases. The coefficients of a logistic regression cannot be directly interpreted as changes in probability, so to estimate the counterfactual effect of incarceration, we use each regression model to predict each individual's baseline probability of recidivism, then use it again to predict each individual's probability of recidivism if they had served one additional year in prison. These results are displayed in Table \ref{tab:logpred}. For the complete population, we see that an additional year of recidivism increases the probability of recidivism by an average of 4.5\%, and that the result remains positive when looking only at paroled individuals. Interestingly, the detrimental effect of incarceration is much stronger for individuals whose most recent offense was a felony rather than a misdemeanor, meaning that additional prison time is particularly harmful for those whose most recent crime was more severe.

As discussed in Section 2.2, we also conduct a Cox proportional hazards regression, with recidivism as a time-to-event endpoint. As seen in Table \ref{tab:NCsample}, the North Carolina dataset contains information on follow-up time for each individual ($t_2$, which we interpret as censoring time) and time of return to prison in the event of recidivism ($t_3$, conditional on $y=1$, which we interpret as event time). Table \ref{tab:survival} presents results of our survival analysis, with the same adjustment set used previously.

We continue to observe a positive treatment effect (the coefficient of prison time) in all four cases. The coefficients of a survival regression also cannot be directly interpreted, so to estimate the counterfactual effect of incarceration, we use each regression model to predict each individual's baseline probability of recidivism within five years, then use it again to predict each individual's probability of recidivism if they had served one additional year in prison. These results are displayed in Table \ref{tab:survpred}, with similar results to those seen in Table \ref{tab:logpred}. As a benefit of survival analysis in practice, note that the five-year period could be modified to any other interval if desired.

\begin{table}[]
\begin{tabularx}{\linewidth}{@{} L *{4}{C{\widthof{Contraceptive}}} @{}} 
\toprule
 & (1) & (2) & (3) & (4) \\
Variables & All offenders & Parolees only & Felons only & Misd. only \\ 
\midrule
Prison time (years) & 0.20*** & 0.27*** & 0.27*** & 0.11*** \\
 & (0.01) & (0.02) & (0.02) & (0.02) \\ \addlinespace
Alcohol use & 0.27*** & 0.33*** & 0.36*** & 0.20*** \\
 & (0.05) & (0.05) & (0.09) & (0.06) \\ \addlinespace
Prior offenses & 0.09*** & 0.07*** & 0.07*** & 0.08*** \\
 & (0.01) & (0.01) & (0.02) & (0.01) \\ \addlinespace
Property crime & 0.42*** & 0.45*** & 0.37*** & 0.35*** \\
 & (0.05) & (0.06) & (0.10) & (0.07) \\ \addlinespace
Personal crime & -0.18* & -0.22* & -0.48*** & 0.06 \\
 & (0.08) & (0.10) & (0.13) & (0.12) \\ \addlinespace
Felony crime & -0.29*** & -0.35*** & & \\
 & (0.06) & (0.07) & & \\ \addlinespace
Constant & -1.05*** & -1.20*** & -1.47*** & -0.93*** \\
 & (0.04) & (0.05) & (0.10) & (0.05) \\ \addlinespace
Observations & 10,357 & 7,877 & 3,866 & 6,491 \\
\midrule
\multicolumn{5}{c}{Robust standard errors in parentheses; *** $p<0.001$, ** $p<0.01$, * $p<0.05$} \\
\end{tabularx} 
\caption{Logistic regression results for causal graph adjustment set}
\label{tab:logistic}
\end{table}

\begin{table}[]
\begin{tabularx}{\linewidth}{@{} L *{4}{C{\widthof{Contraceptive}}} @{}} 
\toprule
 & (1) & (2) & (3) & (4) \\
Recidivism & All offenders & Parolees only & Felons only & Misd. only \\ 
\midrule
Mean baseline probability & 37.3\% & 36.7\% & 39.2\% & 36.2\% \\ \addlinespace
Mean counterfactual probability when adding one year in prison & 41.8\% & 42.7\% & 45.1\% & 38.7\% \\
Difference & 4.5\% & 6.0\% & 5.9\% & 2.5\% \\ 
\midrule
\end{tabularx} 
\caption{Logistic regression counterfactual outcomes under adjusted model}
\label{tab:logpred}
\end{table}

\begin{table}[]
\begin{tabularx}{\linewidth}{@{} L *{4}{C{\widthof{Contraceptive}}} @{}} 
\toprule
 & (1) & (2) & (3) & (4) \\
Variables & All offenders & Parolees only & Felons only & Misd. only \\ 
\midrule
Prison time (years) & 0.11*** & 0.13*** & 0.14*** & 0.08*** \\
 & (0.01) & (0.01) & (0.01) & (0.01) \\ \addlinespace
Alcohol use & 0.25*** & 0.27*** & 0.33*** & 0.20*** \\
 & (0.04) & (0.04) & (0.06) & (0.04) \\ \addlinespace
Prior offenses & 0.05*** & 0.05*** & 0.05*** & 0.05*** \\
 & (0.01) & (0.01) & (0.01) & (0.01) \\ \addlinespace
Property crime & 0.39*** & 0.41*** & 0.32*** & 0.37*** \\
 & (0.04) & (0.05) & (0.08) & (0.05) \\ \addlinespace
Personal crime & -0.07 & -0.09 & -0.31** & 0.13 \\
 & (0.07) & (0.08) & (0.10) & (0.10) \\ \addlinespace
Felony crime & -0.18*** & -0.16*** & & \\
 & (0.04) & (0.05) & & \\ \addlinespace
Observations & 10,357 & 7,877 & 3,866 & 6,491 \\
\midrule
\multicolumn{5}{c}{Robust standard errors in parentheses; *** $p<0.001$, ** $p<0.01$, * $p<0.05$} \\
\end{tabularx} 
\caption{Cox regression results for causal graph adjustment set}
\label{tab:survival}
\end{table}

\begin{table}[]
\begin{tabularx}{\linewidth}{@{} L *{4}{C{\widthof{Contraceptive}}} @{}} 
\toprule
 & (1) & (2) & (3) & (4) \\
Recidivism \emph{within five years} & All offenders & Parolees only & Felons only & Misd. only \\ 
\midrule
Mean baseline probability & 37.3\% & 36.8\% & 39.2\% & 36.2\% \\ \addlinespace
Mean counterfactual probability when adding one year in prison & 40.6\% & 40.5\% & 43.4\% & 38.4\% \\
Difference & 3.3\% & 3.7\% & 4.2\% & 2.2\% \\ 
\midrule
\end{tabularx} 
\caption{Cox regression counterfactual outcomes under adjusted model}
\label{tab:survpred}
\end{table}

\subsection{Double/Debiased Machine Learning}
\label{sec:dmlresult}

Using the \textbf{DoubleML} package, we can select from several options for our treatment and outcome models. As a flexible baseline model that allows for nonlinear relationships between the covariates, we apply random forests \citep{breiman_random_2001}. To check robustness of results, we also apply a handful of alternative models: gradient boosting, support vector machines (SVM), lasso regression, and elastic net regression (with $\alpha = 0.5$). All models are implemented with default hyperparameters; this was found to have little to no impact on the outcomes. Results are presented in Table \ref{tab:double}.

\begin{table}[]
\begin{tabularx}{\linewidth}{@{} L *{4}{C{\widthof{Contraceptive}}} @{}} 
\toprule
 & (1) & (2) & (3) & (4) \\
Models & All offenders & Parolees only & Felons only & Misd. only \\ 
\midrule
Random forests & \textbf{0.045***} & 0.055*** & 0.059*** & 0.025*** \\
 & \textbf{(0.003)} & (0.004) & (0.004) & (0.005) \\ \addlinespace
Gradient boosting & 0.038*** & 0.047*** & 0.053*** & 0.020 \\
 & (0.006) & (0.007) & (0.008) & (0.011) \\ \addlinespace
Support vector machines & 0.060*** & 0.069*** & 0.061*** & 0.042*** \\
 & (0.003) & (0.004) & (0.004) & (0.005) \\ \addlinespace
%Extreme gradient boosting & 0.016*** & 0.023*** & 0.023*** & 0.001 \\
% & (0.002) & (0.003) & (0.003) & (0.004) \\ \addlinespace
Lasso regression & 0.041*** & 0.050*** & 0.054*** & 0.023*** \\
 & (0.003) & (0.003) & (0.004) & (0.005) \\ \addlinespace
%Ridge regression & 0.040*** & 0.051*** & 0.056*** & 0.023*** \\
% & (0.003) & (0.003) & (0.004) & (0.005) \\ \addlinespace
Elastic net regression ($\alpha = 0.5$) & 0.041*** & 0.051*** & 0.054*** & 0.023*** \\
 & (0.003) & (0.003) & (0.003) & (0.005) \\ \addlinespace
Observations & 10,357 & 7,877 & 3,866 & 6,491 \\
\midrule
\multicolumn{5}{c}{Robust standard errors in parentheses; *** $p<0.001$, ** $p<0.01$, * $p<0.05$} \\
\end{tabularx} 
\caption{Double machine learning results for causal effects}
\label{tab:double}
\end{table}

\begin{figure}
\centering
\includegraphics[width=.6\textwidth]{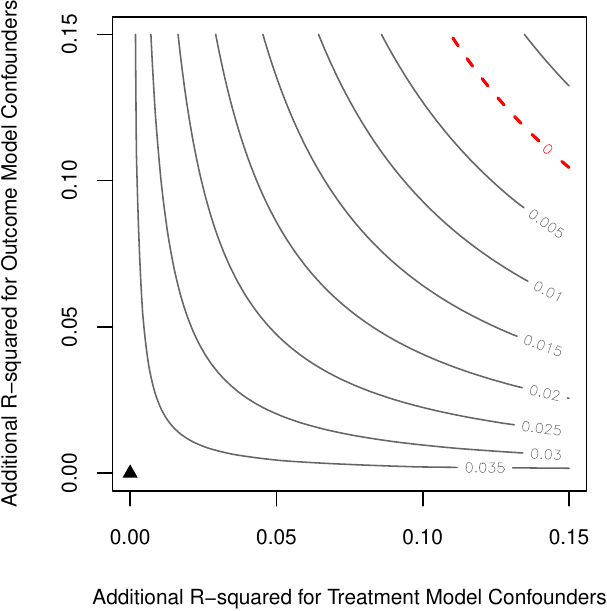}
\caption{Contour plot for double machine learning sensitivity analysis, showing additional explanatory power required for unobserved confounders to affect main causal effect}
\label{fig:sens}
\end{figure}

The results are in line with the previous causal graph analysis: the random forest and other models all indicate that an additional year of prison time increases the probability of future recidivism by about 4\%. This effect is more pronounced for individuals whose most recent offense was a felony rather than a misdemeanor. We note that the choice of model to estimate nuisance parameters within the DML framework can have a marked impact, with the primary causal effect for the full population ranging from 3.8\% to 6\% when using gradient boosting and SVM respectively.

For sensitivity analysis, we focus on the main DML causal effect of 4.5\% when using random forest models, marked in \textbf{bold} in Table \ref{tab:double}. Figure \ref{fig:sens}, developed with the \textbf{dml.sensemakr} package \citep{cinelli2023sensitivity}, displays a contour plot showing the additional explanatory power required for an unobserved confounder to modify this result. We find that any such confounders would need to provide an additional R-squared of 14.1\% for both treatment and outcome to send the estimated effect to zero (12.7\% when including the confidence bound). As discussed in Section \ref{sec:data}, such a variable or set of variables could exist, but this is highly unlikely in light of the broad spectrum of fields that have already been included.

\section{Conclusion}

Modern advances in causal inference provide an array of different methods to analyze observational data, adjust for confounding effects, and identify the treatment effect of a given intervention variable on an outcome of interest. Although prior literature on criminal recidivism focuses almost exclusively on predictive modeling, causal analysis can enable a deeper insight into how specific policy decisions, interventions, or procedures affect recidivism, both for the population at large and for specific subgroups of interest. 

In this case study, using directed acyclic graph (DAG) adjustment and double machine learning (DML), we consistently find that for our specific setting (North Carolina in 1978 and 1980), an additional year in prison caused the probability of recidivism to increase by 3\% to 5\%. This effect was larger for individuals whose most recent offense was a felony, ranging from 4\% to 6\%, suggesting that the marginal impact of incarceration may vary based on the severity of prior offenses. We again emphasize that these results should not be interpreted more broadly, but hope that this paper can serve as a guide for more nuanced causal analysis of criminal justice and policy questions.

% Acknowledgements and Disclosure of Funding should go at the end, before appendices and references

\acks{We thank Robin Evans and Lara Hankeln for their helpful revisions and feedback. We also thank Carlos Carvalho and Rebecca Wilcox for guidance on an earlier version of this work. VS gratefully acknowledges support from the EPSRC Centre for Doctoral Training in Modern Statistics and Statistical Machine Learning (EP/S023151/1) and Novo Nordisk. The opinions and statements expressed in this paper are solely the work of the authors, and do not necessarily reflect the views or positions of any other related entities.}


\newpage

\appendix

\section{}
\label{app:NCdata}

We reprint Table \ref{tab:NCsample} here for convenience. The description and mean value of each column are displayed in Table \ref{tab:NCdetails}. Complete data and code, along with the original data dictionary from \citet{schmidt_data_1989}, can be found at \url{https://github.com/vshirvaikar/recidivism}.

\begin{table}[h]
\centering
\begin{tabular}{ccccccccccccccccc}
\toprule
& $y$ & $t_1$ & $t_2$ & $t_3$ & $x_1$ & $x_2$ & $x_3$ & $x_4$ & $x_5$ & $x_6$ & $x_7$ & $x_8$ & $x_9$ & $x_{10}$ & $x_{11}$ & $x_{12}$ \\
\midrule
Name & \parbox[t]{2mm}{{\rotatebox[origin=c]{90}{Recidivism}}} & 
\parbox[t]{2mm}{{\rotatebox[origin=c]{90}{Prison Time}}} & 
\parbox[t]{2mm}{{\rotatebox[origin=c]{90}{Follow-Up Time}}} & 
\parbox[t]{2mm}{{\rotatebox[origin=c]{90}{Return Time}}} &
\parbox[t]{2mm}{{\rotatebox[origin=c]{90}{Age}}} & 
\parbox[t]{2mm}{{\rotatebox[origin=c]{90}{Gender}}} & 
\parbox[t]{2mm}{{\rotatebox[origin=c]{90}{Race}}} &
\parbox[t]{2mm}{{\rotatebox[origin=c]{90}{Marital Status}}} & 
\parbox[t]{2mm}{{\rotatebox[origin=c]{90}{Felony Crime}}} & 
\parbox[t]{2mm}{{\rotatebox[origin=c]{90}{  Property Crime  }}} & 
\parbox[t]{2mm}{{\rotatebox[origin=c]{90}{  Personal Crime  }}} & 
\parbox[t]{2mm}{{\rotatebox[origin=c]{90}{Prior Offenses}}} & 
\parbox[t]{2mm}{{\rotatebox[origin=c]{90}{Parole Status}}} & 
\parbox[t]{2mm}{{\rotatebox[origin=c]{90}{Education}}} & 
\parbox[t]{2mm}{{\rotatebox[origin=c]{90}{Alcohol Use}}} & 
\parbox[t]{2mm}{{\rotatebox[origin=c]{90}{Hard Drug Use}}} \\
\midrule
A & 0 & 2.5 & 6 & N/A & 36.8 & 1 & 1 & 1 & 0 & 0 & 0 & 0 & 1 & 7 & 1 & 0 \\
B & 1 & 0.6 & 6.8 & 2.4 & 24.3 & 1 & 0 & 1 & 0 & 1 & 0 & 8 & 0 & 9 & 0 & 0 \\
C & 0 & 1.1 & 5.9 & N/A & 23.1 & 1 & 1 & 0 & 0 & 0 & 0 & 1 & 1 & 12 & 0 & 0 \\
\bottomrule
\end{tabular}
\caption{North Carolina dataset sample view (reprint of Table 1)}
\end{table}

\begin{table}[h]
\centering
\begin{tabular}{cccc}
\toprule
& Field & Notes & Mean Value \\
\midrule
$y$ & Recidivism & Binary, within follow-up period $t_2$ of 4-6 years & 0.51 \\
$t_1$ & Prison Time & Total time served in years & 1.61 \\
$t_2$ & Follow-Up Time & Time after release in years when $y$ was recorded & 5.16 \\
$t_3$ & Return Time & Time in years of subsequent offense (if $y=1$) & 1.77 \\
$x_1$ & Age & Age in years & 28.52 \\
$x_2$ & Gender & Binary, male = 1 & 0.94 \\
$x_3$ & Race & Binary, non-black = 1 & 0.51 \\
$x_4$ & Marital Status & Binary, married = 1 & 0.25 \\
$x_5$ & Felony Crime & Binary, crime degree indicator & 0.37 \\
$x_6$ & Property Crime & Binary, crime type indicator & 0.36 \\
$x_7$ & Personal Crime & Binary, crime type indicator & 0.09 \\
$x_8$ & Prior Offenses & Number of previous incarcerations & 1.36 \\
$x_9$ & Parole Status & Binary, supervised release indicator & 0.76 \\
$x_{10}$ & Education & Formal schooling in years & 9.65 \\
$x_{11}$ & Alcohol Use & Binary, ``serious problem'' = 1 & 0.29 \\
$x_{12}$ & Hard Drug Use & Binary, any hard drug use = 1 & 0.23 \\
\bottomrule
\end{tabular}
\caption{North Carolina dataset details}
\label{tab:NCdetails}
\end{table}

\newpage
\bibliography{export-data}

\end{document}